\documentclass[10pt,twocolumn]{article}

\usepackage{iccv}
\usepackage{times}
\usepackage{epsfig}
\usepackage{graphicx}
\usepackage{amsmath}
\usepackage{amssymb}
\usepackage{booktabs}
\usepackage{multirow}
\usepackage{bbding}
\usepackage{comment}
\usepackage{subfigure}
\usepackage{times}
\usepackage{array}
\usepackage{latexsym}
\usepackage{bm}

\usepackage[ruled,vlined]{algorithm2e}%[ruled,vlined]{
\usepackage{algpseudocode}
\renewcommand{\algorithmicrequire}{\textbf{Input:}} 
\renewcommand{\algorithmicensure}{\textbf{Output:}}

\usepackage[pagebackref=true,breaklinks=true, colorlinks, bookmarks=false, hyperfootnotes=false]{hyperref}

\iccvfinalcopy % *** Uncomment this line for the final submission

\def\iccvPaperID{3021} % *** Enter the ICCV Paper ID here

\ificcvfinal\fi

\begin{document}

%%%%%%%%% TITLE
\title{Towards Real-world X-ray Security Inspection: A High-Quality Benchmark And Lateral Inhibition Module For Prohibited Items Detection}% }

\author{
	Renshuai Tao$^{1, 2}$ \quad
	Yanlu Wei$^1$\quad
	Xiangjian Jiang$^{1}$ \quad
	Hainan Li$^{1}$\\
	Haotong Qin$^{1}$ \quad
	Jiakai Wang$^{1}$ \quad
	Yuqing Ma$^1$ \quad
	Libo Zhang$^3$ \quad
	Xianglong Liu$^1$\thanks{corresponding author}\\
	$^1$State Key Laboratory of Software Development Environment, Beihang University\\
	$^2$iFLYTEK Research \qquad
	$^3$Institute of Software Chinese Academy of Sciences\\
	{\tt\small \{rstao, weiyanlu, hainan, qinhaotong, jk\_buaa\_scse, mayuqing, xlliu\}@buaa.edu.cn}\\
	{\tt\small silencejiang12138@gmail.com,  libo@iscas.ac.cn}
}

\maketitle
\ificcvfinal\thispagestyle{empty}\fi

%%%%%%%%% ABSTRACT
\begin{abstract}
	Prohibited items detection in X-ray images often plays an important role in protecting public safety, which often deals with color-monotonous and luster-insufficient objects, resulting in unsatisfactory performance. Till now, there have been rare studies touching this topic due to the lack of specialized high-quality datasets. In this work, we first present a \textbf{Hi}gh-quality \textbf{X-ray} (HiXray) security inspection image dataset, which contains \textbf{102,928} common prohibited items of \textbf{8} categories. It is the largest dataset of high quality for prohibited items detection, gathered from the real-world airport security inspection and annotated by professional security inspectors. Besides, for accurate prohibited item detection, we further propose the \textbf{L}ateral \textbf{I}nhibition \textbf{M}odule (LIM) inspired by the fact that humans recognize these items by ignoring irrelevant information and focusing on identifiable characteristics, especially when objects are overlapped with each other.
	Specifically, LIM, the elaborately designed flexible additional module, suppresses the noisy information flowing maximumly by the Bidirectional Propagation (BP) module and activates the most identifiable charismatic, boundary, from four directions by Boundary Activation (BA) module. We evaluate our method extensively on HiXray and OPIXray and the results demonstrate that it outperforms SOTA detection methods.\footnote{The HiXray dataset and the code of LIM are released at \url{https://github.com/HiXray-author/HiXray}.}
\end{abstract}

\begin{figure}[tp!]
	\begin{center}
		\includegraphics[width=\linewidth]{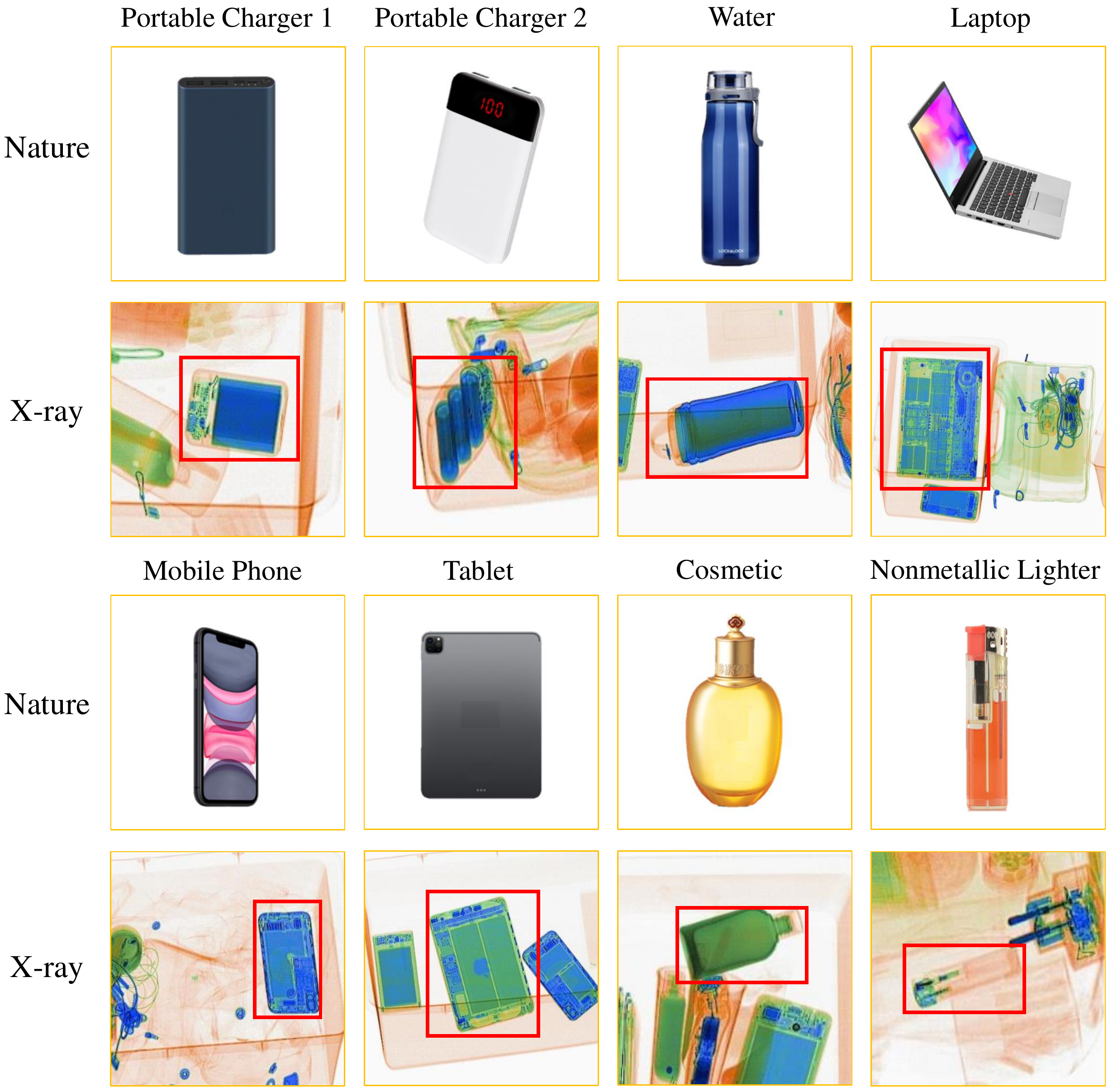}
	\end{center}
	\caption{Nature and X-ray images of the 8 categories of common prohibited items in HiXray. These prohibited items usually contain lithium battery, liquid, lighter, \etc.}
	\label{fig:Category}
\end{figure}

%%%%%%%%% BODY TEXT
\section{Introduction}
As the density if the crowd density increases in public transportation hubs, security inspection has become more and more important in protecting public safety. X-ray scanners, which are adopted usually to scan the luggage and generate the complex X-ray images, play an important role in security inspection scenario. However, security inspectors struggle to accurately detect the prohibited items after a long time highly-concentrating work, which may cause severe danger to the public. Therefore, it is imperative to develop a rapid, accurate and automatic detection method.

Fortunately, the innovation of deep learning \cite{qin2020bipointnet,Qin_2020_CVPR,dualCVPR2021,DBLP:journals/corr/abs-2005-09257,zhang2018hcic,li2021multi}, especially the convolutional neural network, makes it possible to accomplish this goal by transferring it into object detection task in computer vision \cite{lecun2015deep, nah2017deep, young2018recent, hua2018pointwise}. However, different from traditional detection tasks, in this scenario, items within a luggage are randomly overlapped where most areas of objects are occluded resulting in heavy noise in X-ray images. Thus, this characteristic leads to strong requirement of high-quality datasets and models with satisfactory performance for this task.

Regarding dataset, to the best of our knowledge, there are only three released X-ray benchmarks, namely GDXray \cite{mery2015gdxray}, SIXray \cite{miao2019sixray} and OPIXray \cite{WeiOccluded2020}. Both GDXray and SIXray are constructed for classification task and the images of OPIXray are synthetic. Besides, the categories and quantities of labeled instances in the three datasets are far from meeting the requirements in real-world applications. We make a detailed comparison in Table \ref{dataset-comparison}.
Regarding models, traditional CNN-based models \cite{xiao2018deep,fu2018refinet,qin2020forward} trained through common detection datasets fail to achieve satisfactory performance in this scenario because that different from natural images \cite{cheng2017focusing, shi2017detecting} with simple visual information, X-ray images \cite{uroukov2015preLIMinary, mery2016modern, mery2017logarithmic} are characterized by the lacking of strong identification properties and containing heavy noises. This urgently requires researchers to make breakthroughs in both datasets and models.

To address the above drawbacks, in this work, we contribute the largest high-quality dataset for prohibited items detection in X-ray images, named \textbf{Hi}gh-quality \textbf{X-ray} (HiXray) dataset, which contains 102,928 common labeled instances of 8 categories, such as lithium battery, liquid, \etc.
%including ``Portable Charger 1 (lithium-ion prismatic cell)'', ``Portable Charger 2 (lithium-ion cylindrical cell)'', ``Water'', ``Laptop'', ``Mobile Phone'', ``Tablet'', ``Cosmetic'' and     ``Nonmetallic Lighter'' (PO1, PO2, WA, LA, MP, TA, CO and NL, for short). 
All of these images are gathered from real-world daily security inspections in an international airport. Thus, The categories, quantities and locations of prohibited items are in line with the data distribution in real-world scenarios. Besides, each instance is manually annotated by professional inspectors from the international airport, guaranteeing the accurate annotations. In addition, our HiXray dataset can serve the evaluation of various detection tasks including small, occluded object detection, \etc.

For accurate prohibited items detection, we present the \textbf{L}ateral \textbf{I}nhibition \textbf{M}odule (LIM), which is inspired by the fact that humans recognize these items by ignoring irrelevant information and focusing on identifiable characteristics, especially when objects are overlapped with each other. LIM consists of two core sub-modules, namely Bidirectional Propagation (BP) and Boundary Activation (BA). BP filters the noise information to suppress the influence from the neighbor regions to the object regions and BA activates the boundary information as the identification property, respectively. Specifically, BP eliminates noises adaptively through the bidirectional information flowing across layers and BA captures the boundary from four directions inside each layer and aggregates them into a whole outline. 

HiXray dataset and LIM model provide a new and reasonable evaluation benchmark for the community, and helps make a wider breadth of real-world applications. The main contributions of this work are as follows:
\begin{itemize}
	\item We present the largest high-quality dataset named HiXray for X-ray prohibited items detection, providing a new and reasonable evaluation benchmark for the community. We hope that contributing this dataset can promote the development of this issue.
	\item We propose the LIM model, which exploits the lateral inhibition mechanism to improve the detecting ability for accurate prohibited items detection, inspired by the intimate relationship between deep neural networks and biological neural networks.
	\item We evaluate LIM on the HiXray and OPIXray datasets and the results show that LIM can not only be versatile to SOTA detection methods but also improve the performance of them.
\end{itemize}

\section{Related Work}
\textbf{Prohibited Items Detection in X-ray Images.}
X-ray imaging offers powerful ability in many tasks such as medical image analysis \cite{guo2019improved, chaudhary2019diagnosis, lu2019towards} and security inspection \cite{miao2019sixray, huang2019modeling}. As a matter of fact, obtaining X-ray images is difficult, so rare studies touch security inspection in computer vision due to the lack of specialized high-quality datasets.

\begin{table*}[tp!]
	\begin{center}
		\newcommand{\tabincell}[2]{\begin{tabular}{@{}#1@{}}#2\end{tabular}}
		\setlength{\tabcolsep}{1.5mm}
		{
			\small
			\begin{tabular}{lccccccccc}
				\toprule
				\multirow{2}{*}{Dataset} & \multirow{2}{*}{Year} & \multirow{2}{*}{Category} & \multirow{2}{*}{$N_{p}$} & \multicolumn{3}{c}{Annotation} & \multirow{2}{*}{Color} & \multirow{2}{*}{Task} & \multirow{2}{*}{Data Source}\\
				\cmidrule{5-7}
				& && & Bounding Box & Number & 
				Professional & & &\\
				\midrule
				GDXray \cite{mery2015gdxray} & 2015 & 3 & 8,150  & \footnotesize\Checkmark & 8,150  & \footnotesize\XSolidBrush & Gray-scale & Detection & Unknown\\
				SIXray \cite{miao2019sixray} & 2019 & 6 & 8,929 & \footnotesize\XSolidBrush& \footnotesize\XSolidBrush& \footnotesize\XSolidBrush&RGB & Classification & Subway Station\\
				OPIXray \cite{WeiOccluded2020} & 2020 & 5 & 8,885 & \footnotesize\Checkmark & 8,885   & \footnotesize\Checkmark &RGB & Detection & Artificial Synthesis\\
				\midrule
				\textbf{HiXray} &\textbf{2021}&\textbf{8}&\textbf{45,364} & \footnotesize\Checkmark & \textbf{102,928}   & \footnotesize\Checkmark & RGB& Detection& Airport\\
				\bottomrule
			\end{tabular}
		}
	\end{center}
	\caption{Comparison of existing open-source X-ray datasets. $N_{p}$ refers to the number of images containing prohibited items. In our HiXray dataset, some images contain more than one prohibited item and every prohibited item is located with a bounding-box annotation, causing the number of annotation is greater than $N_{p}$.}
	\label{dataset-comparison}
\end{table*}
Several recent efforts \cite{akcay2017evaluation, mery2015gdxray, akcay2018using, miao2019sixray, liu2019deep, WeiOccluded2020} have been devoted to constructing such datasets. A released benchmark, GDXray \cite{mery2015gdxray} contains 19,407 gray-scale images, part of which contain three categories of prohibited items including gun, shuriken and razor blade. SIXray \cite{miao2019sixray} is a large-scale X-ray dataset which is about 100 times larger than the GDXray dataset but the positive samples are less than 1\% to mimic a similar testing environment and the labels are annotated for classification. Recently, \cite{WeiOccluded2020} proposed the OPIXray dataset, containing 8,885 X-ray images of 5 categories of cutters. The images of OPIXray dataset are artificial synthetic. Other relevant works \cite{akcay2017evaluation, akcay2018using, liu2019deep} have not made their data available to download.

\textbf{Object Detection.}
In computer vision area, object detection is one of important tasks, which underpins a few instance-level recognition tasks and many downstream applications. Here we review some works that is the closest to ours. Most of the CNN-based methods can be summarized into two general approaches: one-stage detectors and two-stage detectors.
Recently one-stage methods have gained much attention over two-stage approaches due to their simpler design and competitive performance. SSD \cite{liu2016ssd} discretizes the output space of bounding boxes into a set of default boxes over different aspect ratios and scales. YOLO \cite{redmon2016you, redmon2017yolo9000, redmon2018yolov3, bochkovskiy2020yolov4, yolov5} is the collection of a series of well-known methods, which values both real-time and accuracy among one-stage detection algorithms. Moreover, FCOS \cite{tian2019fcos} proposes a fully convolutional one-stage object detector to solve object detection in a per-pixel prediction fashion, analogue to other dense prediction problems.

\section{HiXray Dataset} 
As Table \ref{dataset-comparison} illustrates, the existing datasets are less than satisfactory and thus fail to meet the requirements in real-world applications. In this work, we construct a new high-quality dataset for X-ray prohibited items detection. Then we introduce the construction principles, data properties and potential tasks of the proposed HiXray dataset.
\subsection{Construction Principles}
We construct the HiXray dataset in accordance with the following five principles:

\textbf{Realistic Source}. Considering realistic source can make the data more meaningful for research, we gather the images of the HiXray dataset from daily security inspections in an international airport to ensure the authenticity of data.

\textbf{Data Privacy}. We strictly follow the standard de-privacy procedure by deleting private information (name, place, \etc), ensuring that nobody can connect the luggage with owners through the images to guarantee the privacy.

\textbf{Extensive Diversity}. HiXray contains the 8 categories of prohibited items such as lithium battery, liquid, lighter, etc., all of which are frequently seen in daily life.

\textbf{Professional Annotation}. Objects in X-ray images are difficult to be recognized for people without professional training. In HiXray, each instance is manually localized with a box-level annotation by professional security inspectors of the airport, who are very skillful in daily work.

\textbf{Quality Control}. 
We followed the similar quality control procedure of annotation as the famous Pascal VOC \cite{everingham2010pascal}. All inspectors followed the same annotation guidelines including what to annotate, how to annotate bounding, how to treat occlusion, \etc. Besides, the accuracy of each annotation was checked by another inspector, including checking for omitted objects to ensure exhaustive labelling.

\subsection{Data Details}
\textbf{Instances per category}. HiXray contains 45,364 X-ray images, 8 categories of 102,928 common prohibited items. The statistics are shown in Table \ref{table-para}.

\begin{table}[!ht]
	\begin{center}
		\setlength{\tabcolsep}{0.5mm}{
			\small
			\begin{tabular}{lccccccccc}
				\toprule
				Category & PO1 & PO2 & WA & LA & MP & TA & CO & NL  & Total  \\
				\midrule
				\footnotesize{Training} & \footnotesize{9,919} & \footnotesize{6,216} & \footnotesize{2,471} & \footnotesize{8,046} & \footnotesize{43,204} & \footnotesize{3,921} & \footnotesize{7,969} & \footnotesize{706} & \footnotesize{82,452}   \\
				\footnotesize{Testing} & \footnotesize{2,502} & \footnotesize{1,572} & \footnotesize{621} & \footnotesize{1,996} & \footnotesize{10,631} & \footnotesize{997} & \footnotesize{1,980} & \footnotesize{177} & \footnotesize{20,476}  \\
				\midrule
				\footnotesize{Total} & \footnotesize{12,421} & \footnotesize{7,788} & \footnotesize{3,092} & \footnotesize{10,042} & \footnotesize{53,835} & \footnotesize{4,918} & \footnotesize{9,949} & \footnotesize{883} & \footnotesize{102,928}   \\
				\bottomrule
			\end{tabular}
		}
	\end{center}
	\caption{The statistics of category distribution of HiXray dataset, where PO1, PO2, WA, LA, MP, TA, CO and NL denote ``Portable Charger 1 (lithium-ion prismatic cell)'', ``Portable Charger 2 (lithium-ion cylindrical cell)'', ``Water'', ``Laptop'', ``Mobile Phone'', ``Tablet'', ``Cosmetic'' and ``Nonmetallic Lighter''.}
	\label{table-para}
\end{table}

\textbf{Instances per image}. On average there are 2.27 instances per image in HiXray dataset. In comparison SIXray has 1.37 (in positive samples), both OPIXray and GDXray have 1 instance per image on average. Obviously, the larger average number of instances per image brings more contextual information, which is more valuable. The statistics is shown in Table \ref{table-para_items}.

\begin{table}[!ht]
	\begin{center}
		\setlength{\tabcolsep}{0.9mm}{
			\small
			\begin{tabular}{lcccccccccc}
				\toprule
				$N_{i}$ & 1 & 2 & 3 & 4 & 5 & 6 & 7 & 8  & 9 &10 \\
				\midrule
				\footnotesize{Training} & \footnotesize{12,726} & \footnotesize{10,905} & \footnotesize{6,860} & \footnotesize{3,286} & \footnotesize{1,521} & \footnotesize{602} & \footnotesize{254} & \footnotesize{91} & \footnotesize{35} &\footnotesize{11}  \\
				\footnotesize{Testing} & \footnotesize{3,227} & \footnotesize{2,722} & \footnotesize{1,705} & \footnotesize{810} & \footnotesize{354} & \footnotesize{145} & \footnotesize{54} & \footnotesize{41} & \footnotesize{8} &\footnotesize{2} \\
				\midrule
				\footnotesize{Total} & \footnotesize{15,953} & \footnotesize{13,627} & \footnotesize{8,565} & \footnotesize{4,096} & \footnotesize{1,875} & \footnotesize{747} & \footnotesize{308} & \footnotesize{132} & \footnotesize{43} &\footnotesize{13}  \\
				\bottomrule
			\end{tabular}
		}
	\end{center}
	\caption{The quantity distribution of images containing different numbers of prohibited items. Note that $N_{i}$ refers to the number of prohibited items in each image.}
	\label{table-para_items}
\end{table}

\textbf{Division of training and testing}. The dataset is partitioned into a training set and a testing set, where the ratio is about 4 : 1. The statistics of category distribution of training set and testing set are also shown in Table \ref{table-para}.

\textbf{Color Information}. Different X-ray machine models may have some differences in color imaging, and we adopt one of the most classic color imaging strategy. The colors of objects under X-ray are mainly determined by their chemical composition, which is introduced in detail in Table \ref{X-ray-color}. %(The model of the security X-ray machine is FISCAN CMEX-T10080, which is widely used in the present.)

\begin{table}[!ht]
	\begin{center}
	\setlength{\tabcolsep}{4.1mm}
	\small
	\begin{tabular}{ccllllc}
		\toprule
		Color  & \multicolumn{5}{c}{Material}             & Typical examples \\ \midrule
		Orange & \multicolumn{5}{c}{Organic Substances}   & Plastics, Clothes   \\
		Blue   & \multicolumn{5}{c}{Inorganic Substances} & Irons, Coppers     \\
		Green  & \multicolumn{5}{c}{Mixtures}             & Edge of phones   \\ \bottomrule
	\end{tabular}
	\end{center}
	\caption{The color information of different objects under X-ray.}
	\label{X-ray-color}
\end{table}

\textbf{Data Quality}. All images are stored in JPG format with a 1200*900 resolution, averagely. The maximum resolution of samples can reach 2000*1040.

\subsection{Potential Tasks}
Our HiXray dataset can further serve the evaluation of various detection tasks including small object detection, occluded object detection, \etc.%and few-shot object detection.

\textbf{Small Object Detection}. Security inspectors often struggle to find small prohibited items in baggage or suitcase. In our HiXray dataset, there are many small prohibited items. According to the definition of small by SPIE, the size of small object is usually no more than 0.12\% of entire image size. We thus define the small object as the object whose ground-truth bounding box accounts for less than 0.1\% of entire image, while the large object is defined as the object whose ground-truth bounding box takes up more than 0.2\% proportion in the entire image, and the rest is the medium. The images of ``Portable Charger 2'' and ``Mobile Phone'' can be divided into three subsets respectively. The categories distribution is illustrated in Table \ref{table-num}.

\begin{table}[!ht]
	\begin{center}
		\setlength{\tabcolsep}{3.7mm}
		\small
		{
			\begin{tabular}{lcccc}
				\toprule
				Category & Total & Large & Medium & Small  \\
				\midrule
				PO2  &2,502 & 587 & 986  & 929  \\
				\midrule
				MP & 10,631 & 3,547 & 4,248 & 2,836  \\
				\bottomrule
			\end{tabular}
		}
	\end{center}
	\caption{The category distribution of ``Portable Charger 2'' and ``Mobile Phone'' (PO2 and MB for short) when the two thresholds are set as 0.1\% and 0.2\%.}
	\label{table-num}
\end{table}

\begin{figure}[!ht]
	\begin{center}
		\includegraphics[width=0.9\linewidth]{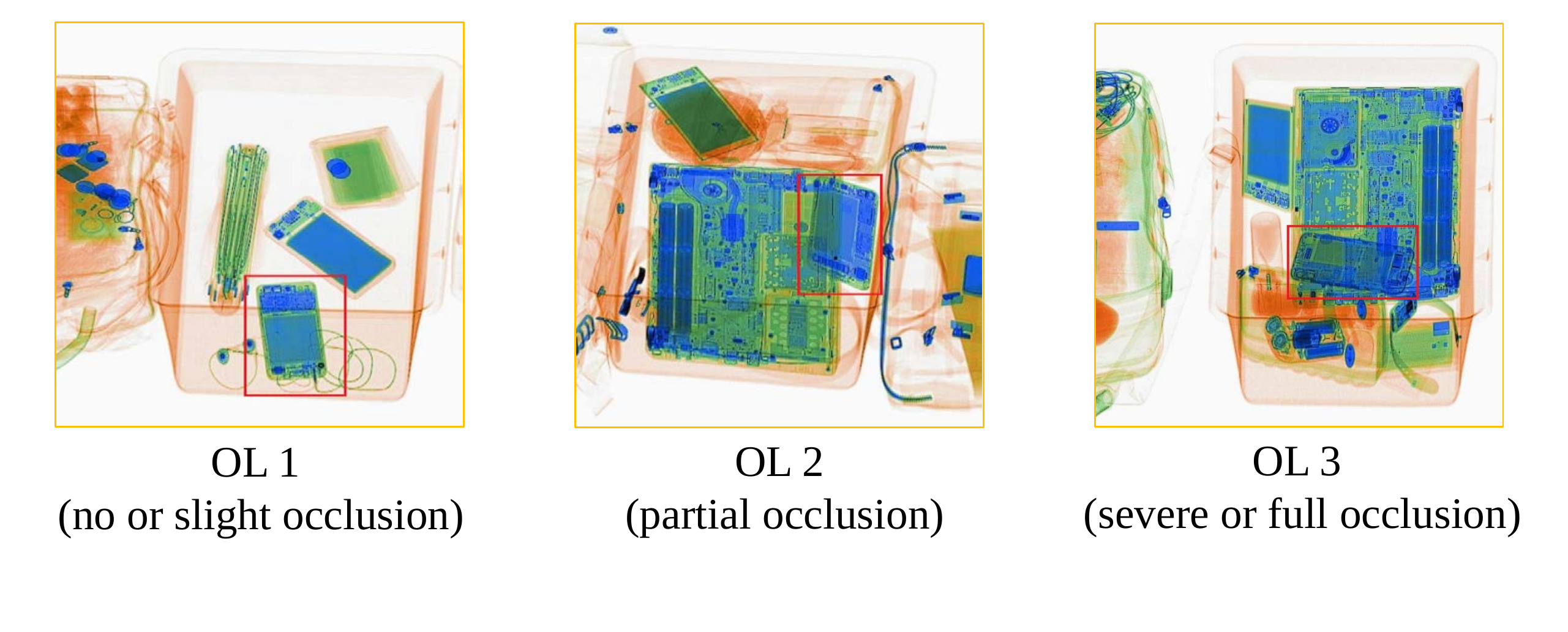}
	\end{center}
	\caption{Three occlusion levels on prohibited items in the images of HiXray, where the numbers indicates occlusion levels.}
	\label{fig:zhedang}
\end{figure}

\textbf{Occluded Object Detection}. The items in baggage or suitcase are often overlapped with each other, causing the occlusion problem in X-ray prohibited items detection. \cite{WeiOccluded2020} proposed the occluded prohibited items detection task in X-ray security inspection. The occlusion problem also exists in HiXray dataset with large-scale images with more categories and numbers. In order to study the impact brought by object occlusion levels, researchers can divide the HiXray dataset into three (or more) subsets according to different occlusion levels (illustrated in Figure \ref{fig:zhedang}).

\begin{figure*}[!ht]
	\begin{center}
		\includegraphics[width=\linewidth]{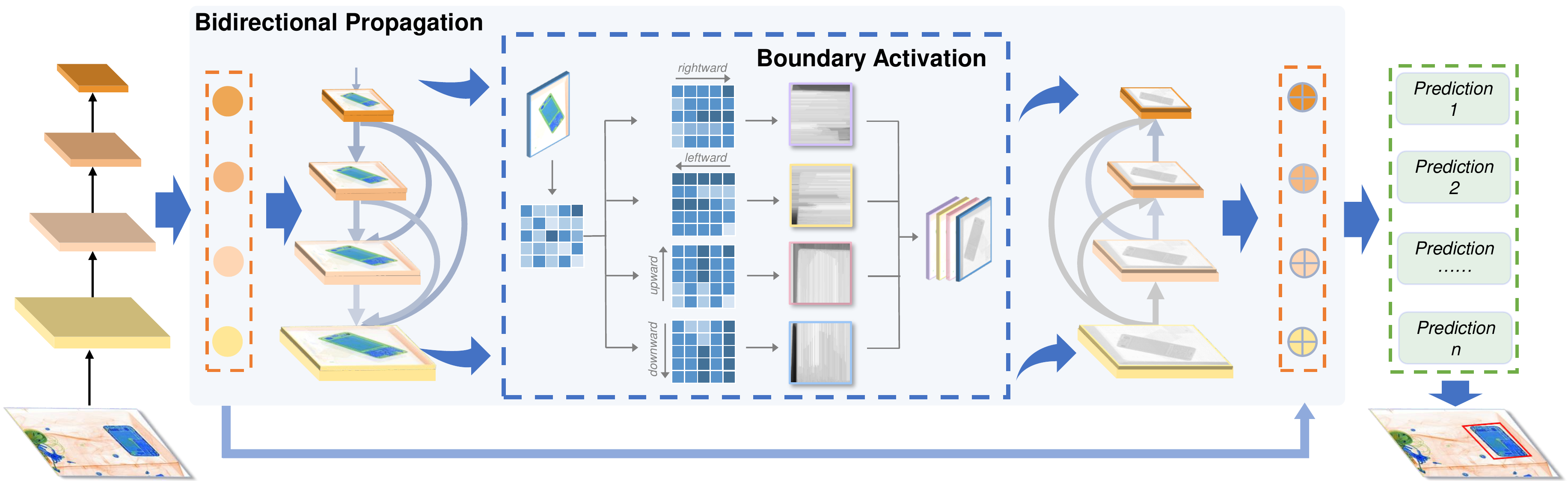}
	\end{center}
	\caption{The network structure of Lateral Inhibition Module (LIM). Bidirectional Propagation filters noisy information to suppress the influence from neighbor regions to object regions and Boundary Activation activates the boundary as the identification property, respectively.}
	\label{fig:structure}
\end{figure*}

\section{The Lateral Inhibition Module}
In neurobiology, lateral inhibition disables the spreading of action potentials from excited neurons to neighboring neurons in the lateral direction. We mimic this mechanism by designing a bidirectional propagation architecture to adaptively filter the noisy information generated by the neighboring regions of the prohibited items. Also, lateral inhibition creates contrast in stimulation that allows increased sensory perception, so we activate the boundary information by intensifying it from four directions inside each layer and aggregating them into a whole shape.

Therefore, inspired by the mechanism that lateral suppression by neighboring neurons in the same layer making the network more efficient, we propose the Lateral Inhibition Module (LIM). In this section, we will first introduce the network architecture in Section \ref{sec:arch} and further the two core sub-modules, namely Bidirectional Propagation (BP) and Boundary Activation (BA), in Section \ref{sec:BA} and Section \ref{sec:BP}, respectively.

\subsection{Network Architecture}\label{sec:arch}

Figure \ref{fig:structure} illustrates the architecture of our LIM.
It takes a single-scale image of an arbitrary size as input, and outputs proportionally sized feature maps at multiple levels. Similar to FPN \cite{lin2017feature} and some other varieties like PANet \cite{wang2019panet}, this process is independent of the backbone architectures.

Specifically, suppose there are $N$ training images $\textbf{X} = \left\{\textbf{x}_1,\cdots,\textbf{x}_N\right\}$ and $L$ convolutional layers in the backbone network. One sample $\textbf{x}\in\textbf{X}$ is fed into the backbone network and computed feed-forwardly, which computes a feature hierarchy consisting of feature maps at several scales with a scaling step of 2.

\renewcommand{\algorithmicrequire}{\textbf{Input:}} 
\renewcommand{\algorithmicensure}{\textbf{Output:}}
\begin{algorithm}[!t]
	\caption{The Procedure of LIM.}
	\label{alg:1}
	\KwIn{The feature map set $\textbf{F}=\{\mathcal{F}^{1}(\textbf{x}),\cdots,\mathcal{F}^{L}(\textbf{x})\}$.}
	\KwOut{The refined feature map set $\textbf{C}=\{\mathbf{C}^{1},\cdots,\mathbf{C}^{L}\}$.}
	\For{all $l=1,2,\dots, L$}
	{
		Compute $\textbf{A}^l$ based on Eq. $\left(\ref{U_1}\right)$\;
		\For{direction in fourDirections}
		{
			\If{direction is horizontal}
			{
				// Avoid column loop for faster speed\;
				Rotate feature map\;
			}
			\For{row $\leftarrow$ 1 to heightOfMap}
			{
				Compute $\textbf{B}^l_{ijc}$ based on Eq. $\left(\ref{BA_formular}\right)$\;
			}
		}
		Generate $\textbf{B}^l$ by concatenating all $\textbf{B}^l_{ijc}$\;
		Compute $\textbf{C}_{\mathrm{t}}^{l}$ based on Eq. $\left(\ref{C_1}\right)$\;
		Compute $\textbf{C}^l$ based on Eq. $\left(\ref{C_2}\right)$\;
	}
	Obtain the feature map set $\textbf{C}$=$\{\mathbf{C}^{1},\cdots,\mathbf{C}^{L}\}$.
	\label{Alg}
\end{algorithm}

Suppose $\mathcal{F}(\cdot)$ as a composite function of three consecutive operations: Batch Normalization (BN) \cite{ioffe2015batch}, followed by a rectified linear unit (ReLU) \cite{glorot2011deep} and a 3 $\times$ 3 Convolution (Conv). $\mathcal{F}^{l}(\textbf{x})$ is the feature map generated by the $l$-th layer of the network. \textbf{Firstly}, in the left part of BP, the noisy information is adaptively filtered because their propagation is reduced from high-level to low-level feature maps in the top-down pathway. \textbf{Secondly}, the output feature maps are fed into BA. BA refines the feature maps to activate the boundary information by enhance it from four directions and outputs the refined feature maps. \textbf{Thirdly}, similar to the left, the right part of BP reduces the propagation of noisy information from low-level to high-level feature maps through the bottom-up pathway. \textbf{Finally}, the feature maps outputted by each layer of the right of BP combine those of the corresponding layer from the backbone network. And the combined feature maps is conveyed to the following prediction layers. Algorithm \ref{Alg} summarizes the whole process (We add explanations about acceleration operation in code implementation in Algorithm \ref{Alg}) and the details of modules are described in the following sections.

\subsection{Bidirectional Propagation}\label{sec:BP}
To disable the spreading of noisy information of neighboring regions, we mimic this mechanism by designing the bidirectional propagation architecture. Moreover, we add a dense mechanism to enhance the ability of BP to choose proper information to propagate.

As shown in Figure \ref{fig:structure}, for the dense top-down pathway on the left of BP, up-sampling spatially coarser but semantically stronger feature maps from higher pyramid levels hallucinates higher resolution features. These feature maps are enhanced by the corresponding feature maps from the convolutional layers via lateral connections. Each lateral connection merges feature maps of the same spatial size from the convolutional layer and the top-down pathway. The feature map of low convolutional layer is of lower-level semantics, but its activation is more accurately localized as it was sub-sampled fewer times. Further, we construct the dense connections to ensure maximum the filter.

Specifically, to preserve the feed-forward nature, $\mathcal{F}^{l}(\textbf{x})$ obtains additional inputs from the feature maps $\mathcal{F}^{l+1}(\textbf{x})$, $\cdots$, $\mathcal{F}^{L}(\textbf{x})$ of all preceding layers and passes on its own feature-maps to the feature maps $\mathcal{F}^{l-1}(\textbf{x})$, $\cdots$, $\mathcal{F}^{1}(\textbf{x})$ of all subsequent layers. Figure \ref{fig:structure} illustrates this layout schematically. We define $\mathcal{U}^m(\cdot)$ as the up-sampling operation ($2^m$ times) and $\mathcal{V}(\cdot)$ as a $1\times1$ convolutional layer to reduce channel dimensions. The process is formulated as follows:

\begin{equation}
\textbf{A}^{l}=\mathcal{V}\left(\mathcal{F}^{l}(\textbf{x})\right)+\sum_{m=1}^{L-l}\mathcal{U}^{m}\left(\textbf{A}^{l+m}\right)
\label{U_1},
\end{equation}
where $\textbf{A}^{l}$ refers to the feature map outputted by the $m$-th layer of the left part of BP.

Regarding the right part of BP, as the Figure \ref{fig:structure} illustrated, suppose the input feature map $\textbf{B}^{l}$ refers to the feature map whose boundary has been activated in Eq. $\left(\ref{BA_formular}\right)$ (Boundary Activation will be introduced in the following section). Similar to the previous definition, $\mathcal{D}^m(\cdot)$ is the down-sampling operation ($2^m$ times). This process can be formulated as follows:

\begin{equation}
\textbf{C}_{\mathrm{t}}^{l}=\mathcal{V}\left(\textbf{B}^{l}\right)+\sum_{m=1}^{l-1} \mathcal{D}^{m}\left(\textbf{C}_{\mathrm{t}}^{l-m}\right)
\label{C_1},
\end{equation}
\begin{equation}
\textbf{C}^{l}=\textbf{C}_{\mathrm{t}}^{l}+\mathcal{F}^{l}(\textbf{x})
\label{C_2},
\end{equation}
where $\textbf{C}_{\mathrm{t}}^{l}$ refers to the output of $l$-th layer of the bottom-up pathway and $\textbf{C}^{l}$ refers to the feature map generated of the $l$-th layer of BP. Finally, we convey $\textbf{C}^{l}$, outputted by LIM, to the following prediction layers.

\begin{figure}[tp!]
	\begin{center}
		\includegraphics[width=1.0\linewidth]{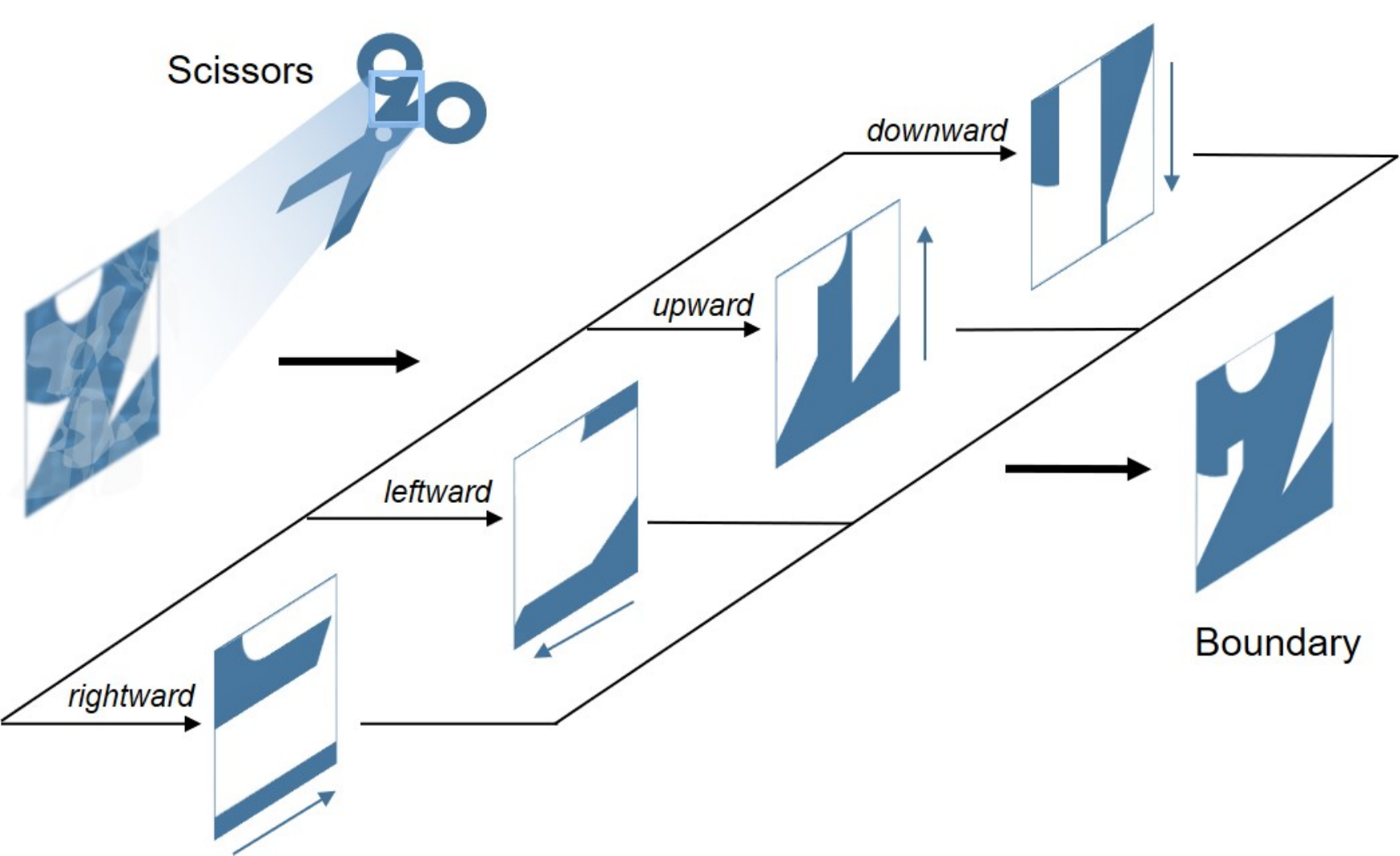}
	\end{center}
	\caption{The schematic diagram of Boundary Activation.}
	\label{fig:theory_and_instance}
\end{figure}

\begin{table*}[!t]
	\begin{center}
		\setlength{\tabcolsep}{2mm}{
			\small
			\begin{tabular}{l|ccccccccc|cccccc}
				\hline
				\multirow{2}{*}{Method} & \multicolumn{9}{c|}{HiXray Dataset (\textbf{Ours})} & \multicolumn{6}{c}{OPIXray Dataset \cite{WeiOccluded2020}}\\
				\cline{2-16}
				& \textbf{AVG}& PO1 & PO2 & WA & LA & MP & TA & CO & NL & \textbf{AVG} & FO & ST & SC & UT & MU  \\
				\hline
				\footnotesize{SSD} \cite{liu2016ssd} & 71.4& 87.3 & 81.0 & 83.0 & 97.6 & 93.5 & 92.2 & 36.1 & 0.01 & 70.9 & 76.9 & 35.0 & 93.4 & 65.9 & 83.3   \\
				\footnotesize{SSD+DOAM} \cite{WeiOccluded2020} & 72.1& 88.6 & 82.9 & 83.6 & 97.5 & 94.1 & 92.1 & 38.2 & 0.01 & 74.0 & 81.4 & 41.5 & 95.1 & 68.2 & \textbf{83.8}   \\
				\footnotesize{\textbf{SSD+LIM}} & \textbf{73.1} &\textbf{89.1}&\textbf{84.3}&\textbf{84.0} & \textbf{97.7} & \textbf{94.5} & \textbf{92.4} & \textbf{42.3} & \textbf{0.1} & \textbf{74.6} &\textbf{81.4}&\textbf{42.4} & \textbf{95.9} & \textbf{71.2} & 82.1  \\
				\hline
				\footnotesize{FCOS} \cite{tian2019fcos} & 75.7& 88.6 & 86.4 & 86.8 & 89.9 & 88.9 & 88.9 & 63.0 & 13.3  & 82.0& 86.4 & 68.5 & 90.2 & 78.4 & 86.6  \\
				\footnotesize{FCOS+DOAM} \cite{WeiOccluded2020} & 76.2 & 88.6& 87.5 & 87.8 & 89.9 & 89.7 & 88.8 & 63.5 & 12.7 & 82.4 & 86.5 & 68.6 & 90.2 & 78.8 & \textbf{87.7}  \\
				\footnotesize{\textbf{FCOS+LIM}} & \textbf{77.3} & \textbf{88.9}& \textbf{88.2} & \textbf{88.3} & \textbf{90.0} & \textbf{89.8} & \textbf{89.2} & \textbf{69.8} & \textbf{14.4} & \textbf{83.1} & \textbf{86.6} & \textbf{71.9} & \textbf{90.3} & \textbf{79.9} & 86.8   \\
				\hline
				\footnotesize{YOLOv5} \cite{yolov5} & 81.7& 95.5 & 94.5 & 92.8 & 97.9 & 98.0 & 94.9 & 63.7 & 16.3 & 87.8 & 93.4 & 67.9 & 98.1 & 85.4 & 94.1   \\
				\footnotesize{YOLOv5+DOAM} \cite{WeiOccluded2020} & 82.2& 95.9 & 94.7 & 93.7 & 98.1 & 98.1 & 95.8 & 65.0 & 16.1 & 88.0 & 93.3 & 69.3 & 97.9 & 84.4 & \textbf{95.0}  \\
				\footnotesize{\textbf{YOLOv5+LIM}} & \textbf{83.2}& \textbf{96.1} & \textbf{95.1} & \textbf{93.9} & \textbf{98.2} & \textbf{98.3} & \textbf{96.4} & \textbf{65.8} & \textbf{21.3} & \textbf{90.6}  & \textbf{94.8} & \textbf{77.6} & \textbf{98.2} & \textbf{88.9} & 93.8  \\
				\hline
			\end{tabular}
		}
	\end{center}
	\caption{Comparisons of common detection approaches SSD, FCOS and YOLOv5 (for simplicity, we use the lightest YOLOv5s model in the YOLOv5 experiment), and the latest related model DOAM on the HiXray dataset and OPIXray dataset, where the definition of the categories (PO1, PO2, \etc.) can be found in Table \ref{table-para}. FO, ST, SC, UT and MU donate ``Folding Knife'', ``Straight Knife'', ``Scissor'', ``Utility Knife'' and ``Multi-tool Knife'' in the OPIXray dataset, respectively.
	}
	\label{table-traditional}
\end{table*}
\subsection{Boundary Activation}\label{sec:BA}
To mimic the mechanism that lateral inhibition creates contrast in stimulation that allows increased sensory perception, we activate the boundary information by intensifying it from four directions inside the feature maps outputted by each layer and aggregating them into a whole shape. The schematic diagram is shown in Figure \ref{fig:theory_and_instance}.

As is shown in Figure \ref{fig:theory_and_instance}, the key to capturing the boundary of object is to determine whether a position is a boundary-point. Motivated by the schematic diagram, we design the module Boundary Activation to perceive the sudden changes of boundary and its surroundings. Suppose we want to capture the left boundary of the object in the feature map $\textbf{A}^{l} \in \mathbb{R}^{{H}\times{W}\times{C}}$ (the output of left part of Bidirectional Propagation). $\textbf{A}^{l}_{c}$ donates the $c$-th channel of $\textbf{A}^{l}$. Further, $\textbf{A}^{l}_{ijc}$ refers to the location ($i,j$) of the feature map $\textbf{A}^{l}_{c}$. To determine whether there is a sudden change between a position and the left of the point, the right-most point $\textbf{A}^{l}_{iWc}$ traverses to the left. The process of perceiving the left boundary can be formulated as Eq. $\left(\ref {BA_formular}\right)$.

\begin{equation}
\small
\textbf{B}^{l}_{ijc}\! =\!\left\{\! 
\begin{array}{cc}
\textbf{A}^{l}_{iWc}\quad  &\text{if } j=W, \\
\\
\max \left\{\!\textbf{A}^{l}_{ijc},\textbf{A}^{l}_{i(j+1)c},\dots,\textbf{A}^{l}_{iWc}\right\}\!\!&\! \text {otherwise,}
\end{array}\right.
\label{BA_formular}
\end{equation}
where the $\textbf{B}^{l}_{ijc}$ refers to the location ($i,j$) of $c$-th channel of the feature map $\textbf{B}^{l}$ after Boundary Activation.

% \usepackage{multirow}

\section{Experiments}
In this section, we conduct comprehensive experiments on HiXray and OPIXray dataset to evaluate the effectiveness of LIM. To the best of our knowledge, HiXray and  OPIXray \cite{WeiOccluded2020} are the only two datasets currently available for X-ray prohibited items detection (RGB).

First, we verify the effectiveness of LIM by comparing the base and LIM-integrated classic or SOTA detection methods (SSD \cite{liu2016ssd}, FCOS \cite{tian2019fcos} and YOLOv5 \cite{yolov5}). We evaluate all the base detection methods and the LIM-integrated methods on HiXray and OPIXray datasets. Second, we evaluate the superiority of our LIM over other feature pyramid mechanisms by comparing two famous methods FPN \cite{lin2017feature} and PANet \cite{wang2019panet} on the HiXray dataset.
Third, we perform an ablation study to thoroughly evaluate each part of LIM. Finally, we conduct the visualization experiment to demonstrate the performance improvement.

\subsection{Experiment Setting Details}
\textbf{LIM:} LIM is implemented by PyTorch for its high flexibility and powerful automatic differentiation mechanism. The LIM-integrated model refers to the model that we implement this mechanism inside (Section~\ref{s52}). Both FPN and PANet contain feature pyramid mechanism similar to LIM, but they are not plug-in model. Therefore, we refer to their published code and  re-implemented the mechanism deployed in SSD (Section~\ref{s53}). Unless specified, we use the following implementation details.

\textbf{Backbone Networks:} The backbone networks of SSD, FCOS and YOLOv5 are VGG16 \cite{simonyan2014very}, ResNet50 \cite{he2016deep} and CSPNet \cite{wang2020cspnet} respectively. For each backbone network, we modify the corresponding network architecture to implement LIM mechanism.

\textbf{Parameters:} All experiments of LIM and baselines are optimized by the SGD optimizer and the initial learning rate is set to 0.0001. The momentum and weight decay are set to 0.9 and 0.0005 respectively. The batch size is set to 32 with shuffle strategy while training. We evaluate the mean Average Precision (mAP) of the object detection to measure the performance of all models fairly. Besides, the IOU threshold measuring the accuracy of the predicted bounding box against the ground-truth is set to 0.5.

\subsection{Comparing with SOTA Detection Methods}
\label{s52}

We verify the effectiveness of LIM by implementing this mechanism to several detection approaches, including traditional SSD, the latest FCOS and YOLOv5. We integrate LIM to the three detection approaches and compare the LIM-integrated methods to the original baselines. In addition, we integrate the latest detection method for security inspection DOAM (in the work of OPIXray dataset) into the three detection approaches above and compare the results with our LIM. The experimental results on HiXray dataset and OPIXray dataset are shown in Table \ref{table-traditional}.

Table \ref{table-traditional} demonstrates that in HiXray dataset, the LIM-integrated network improves the average performance by 1.7\%, 1.6\% and 1.5\% over the original base models SSD, FCOS and YOLOv5 respectively. Besides, LIM outperforms DOAM by 1\%, 1.1\% and 1\% with the base model SSD, FCOS and YOLOv5 respectively. In OPIXray dataset, the LIM-integrated network improves the mean performance by 3.7\%, 1.1\% and 2.8\% over the original models SSD, FCOS and YOLOv5 respectively. Besides, LIM outperforms DOAM by 0.6\%, 0.7\% and 2.6\% with the base models SSD, FCOS and YOLOv5 respectively.

Note that the performances are particularly low in two classes (CO and NL) for all models in Table \ref{table-traditional}, it is mainly because that, compared to other categories, NL and CO are far more difficult to recognize. For NL, it is very small in size and composed of a small piece of iron and a plastic body. The plastic shown under X-ray appears orange, which almost blends in with the background. For CO, the main reason is that there are big differences in the shapes of cosmetics, such as round and square, which are easy to be confused with other kinds of items.

\subsection{Comparing with Feature Pyramid Mechanisms}
\label{s53}

LIM can be regarded as another feature pyramid method with a novel dense connection mechanism with specific feature enhancement. Therefore, We compare our LIM with the classical feature pyramid mechanism FPN and the variety PANet in different base models. Note that there is the same feature pyramid mechanism of FPN in FCOS and a variety of PANet mechanism in YOLOv5, so we replace the feature pyramid mechanism with our LIM in FCOS and YOLOv5 to verify that our mechanism works better in the base model FCOS and YOLOv5 (the same as Section~\ref{s52}). The experimental results are shown in Table \ref{table-eBFPN}.

\begin{figure*}[tp!]
	\begin{center}
		%\fbox{\rule{1pt}{1pt} \rule{1pt}{1pt}
		\includegraphics[width=\linewidth]{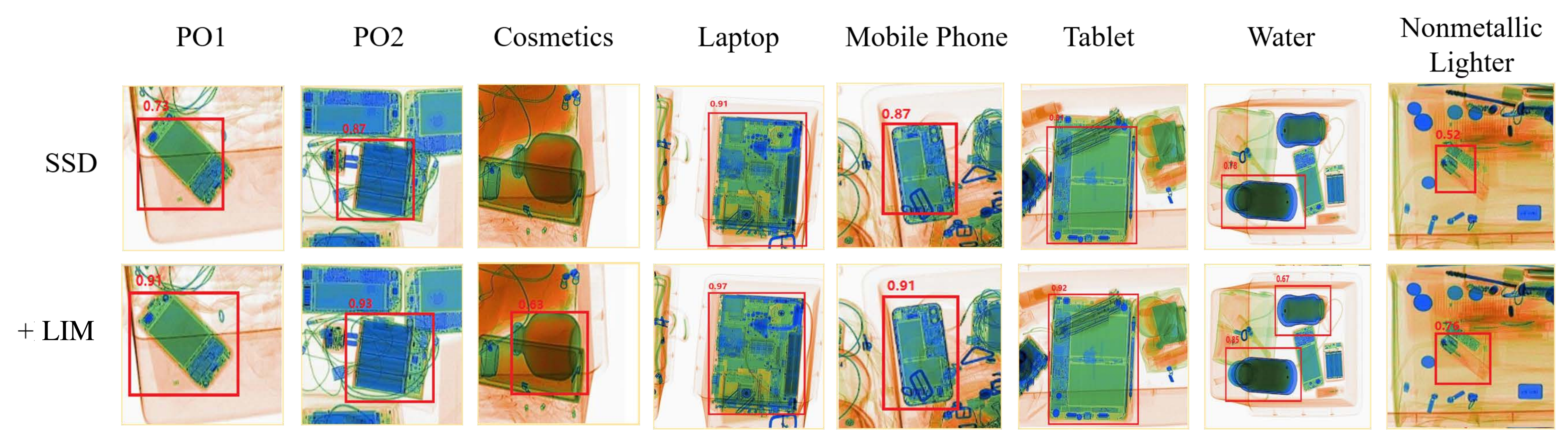}
		%}
	\end{center}
	\caption{Visualization of the performance of both the baseline SSD and the LIM-integrated model. We select one pair of images of each category. The ability to identify and locate prohibited items of the LIM-integrated model outperforms the baseline SSD obviously.}
	\label{fig:contradistinction}
\end{figure*}

\begin{table}[!t]
	\begin{center}
		\setlength{\tabcolsep}{0.7mm}
		{
			\small
			\begin{tabular}{lccccccccc} 
				\toprule
				Method & \textbf{AVG}& PO1 & PO2 & WA & LA & MP & TA & CO & NL  \\
				\midrule
				SSD \cite{liu2016ssd} & 71.4& 87.3 & 81.0 & 83.0 & 97.6 & 93.5 & 92.2 & 36.1 & 0.01  \\
				\midrule
				+FPN \cite{lin2017feature} & 72.0& 87.4 & 81.5 & 83.2 & 97.9 & 93.9 & 92.2 & 40.3 & 0.02  \\
				+PANet \cite{wang2019panet} & 72.0 & 88.3 & 83.2 & 82.8 & \textbf{97.9} & 93.8 & \textbf{92.6}& 37.3 & 0.01 \\
				\textbf{+LIM} & \textbf{73.1}& \textbf{89.1}& \textbf{84.3} &\textbf{84.0}& 97.7 & \textbf{94.5} & 92.4 & \textbf{42.3} & \textbf{0.1}  \\
				\bottomrule
			\end{tabular}
		}
	\end{center}
	\caption{Comparing with feature pyramid mechanisms in the base SSD model, where the feature pyramid mechanisms and our LIM are implemented inside the base model respectively.}    
	\label{table-eBFPN}
\end{table}
LIM improves by both 1.1\% to FPN and PANet in the base SSD model, 1.1\% to FPN in the base FCOS model and 1.5\% to the variety of PANet in the base YOLOv5 model. Further, we observe from the Table \ref{table-eBFPN} that LIM improves significantly than FPN on the categories like ``Portable Charger 1'' (0.7\%), ``Portable Charger 2'' (1.9\%) and ``Water'' (1.1\%). The visual information like boundary of the three categories is more abundant in their X-ray images, demonstrating the effectiveness of Boundary Activation in our LIM and verifies the novel dense connection mechanism with specific feature enhancement.

\subsection{Ablation Study}

In this section, we conduct several ablation studies to in-depth investigate our method. We first analyze the effectiveness of the Dense mechanism by implementing the Single-directional Propagation (the left part of the Boundary Propagation) inside the base model. Then, we evaluate the performance of Boundary Activation alone that no boundary information aggregation inside the feature map. Moreover, we add the Boundary Activation module. The experimental results are shown in Table \ref{table-our}.

In Table \ref{table-our}, we can observe that the performance of the network with only Single-directional Propagation improves by 0.7\% than the base model, verifying the effectiveness of our dense mechanism. After applying the propagation toward another direction, the performance improves by 1.2\% than the base model and 0.5\% than the Single-directional Propagation, which demonstrates the effectiveness of our bidirectional mechanism. Further, Table \ref{table-our} shows that after the integration of our Boundary Activation module, the performance improves 1.7\% than the base model and 0.5\% than Boundary Propagation alone, indicating the effectiveness of boundary information aggregation inside the feature map. In conclusion, ablation studies have verified the validity of each part of our LIM model.

\begin{table}[!t]
	\begin{center}
		\setlength{\tabcolsep}{1mm}
		{
			\small
			\begin{tabular}{lccccccccc}
				\toprule
				Method& \textbf{AVG} & PO1 & PO2 & WA & LA & MP & TA & CO & NL  \\
				\midrule
				SSD \cite{liu2016ssd} & 71.4& 87.3 & 81.0 & 83.0 & 97.6 & 93.5 & 92.2 & 36.1 & 0.01  \\
				\midrule
				+SP & 72.1& 87.9& 82.3 & 83.8& 97.9 & 92.4 & 92.6 & 38.8 & 0.63  \\
				+BP & 72.6& 88.1& 83.4 & 83.9 & \textbf{97.8} & 93.8 & \textbf{92.8} & 40.3 & 0.03  \\
				+BP+BA & \textbf{73.1}&\textbf{89.1}&\textbf{84.3}&\textbf{84.1} & 97.7 & \textbf{94.5} & 92.4 & \textbf{42.3} & \textbf{0.1}  \\
				\bottomrule
			\end{tabular}
		}
	\end{center}
	\caption{Results of ablation study. Note that in this table, SP refers to the single-directional dense propagation (the left part of Boundary Propagation). BP refers to the Boundary Propagation and BA refers to the Boundary Activation.}
	\label{table-our}
\end{table}

\subsection{Visualization}
\begin{figure}[!ht]
	\begin{center}
		\includegraphics[width=\linewidth]{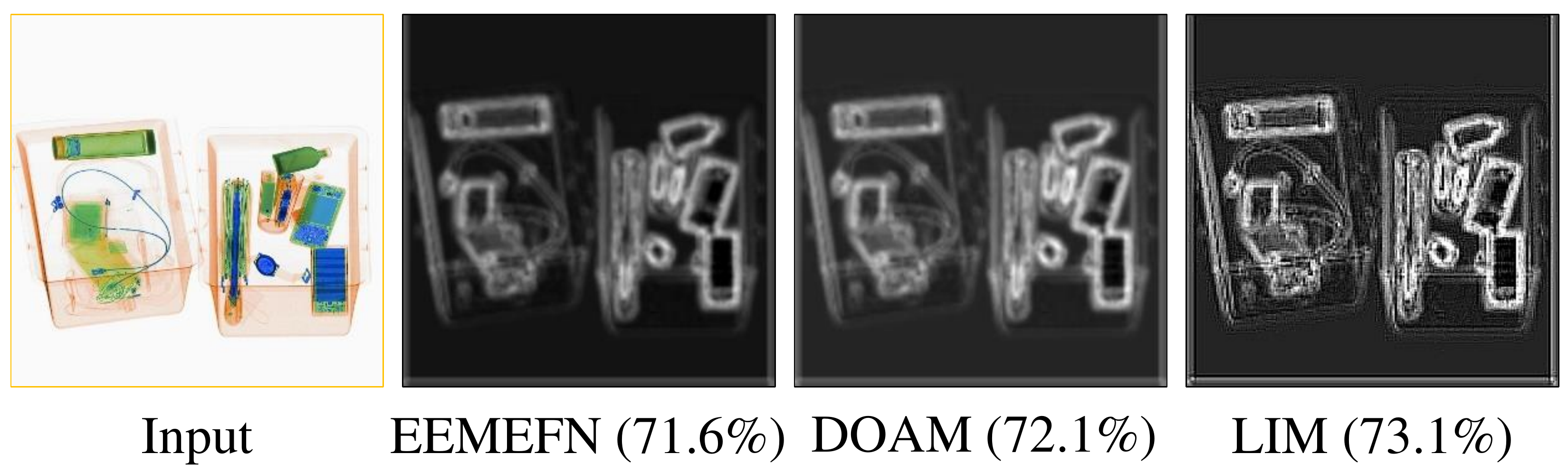}
	\end{center}
	\caption{Visualization of the effectiveness of our LIM.}
	\label{fig:visual_boundary}
\end{figure}

In this section, we visualize the accuracy of recognition and localization in Figure \ref{fig:contradistinction} and the effectiveness of LIM and traditional boundary-enhanced methods in Figure \ref{fig:visual_boundary}.

Figure \ref{fig:contradistinction} shows that the LIM-integrated model has a significant improvement over the baseline. In columns 1, 2, 5, 6 and 8, the detection boundaries of prohibited items by the base SSD model are not precise enough and the LIM-integrated model performs better obviously. In column 3, the cosmetic escapes from the detection of the base SSD model but is caught by LIM-integrated model with the confidence of 91\%. In column 7, the base SSD model only detects one prohibited item while there are two, but both of them are detected by LIM. Figure \ref{fig:visual_boundary} illustrated the effectiveness of our LIM and traditional boundary-enhanced methods, including DOAM \cite{WeiOccluded2020}, EEMEFN \cite{zhu2020eemefn}, \etc.

\section{Conclusion}
In this paper, we investigate prohibited items detection in X-ray security inspection, which plays an important role in protecting public safety. However, this track has not been widely studied due to the lack of specialized public datasets. To facilitate research in this field, we construct and release a dataset with high-quality X-ray images for prohibited items detection, namely HiXray, including 8 categories of 102,928 common prohibited items. All images are gathered from the real-world scenario and manually annotated by professional inspectors. Besides, we propose the Lateral Inhibition Module (LIM) to address the problem that the items to be detected are usually overlapped with the stacked objects during X-ray imaging. Inspired by the lateral suppression mechanism of neurobiology, LIM eliminates the influence of noisy neighboring regions on the object regions of interest and activates the boundary of items by intensifying it.
We comprehensively evaluate LIM on the HiXray and OPIXray dataset and the results demonstrate that LIM can improve the performance of SOTA detection methods. We hope that contributing this high-quality data set and LIM model can promote the rapid development of prohibited items detection in X-ray security inspection.

\section*{Acknowledge}
This work was supported by National Natural Science Foundation of China (62022009, 61872021), Beijing Nova Program of Science and Technology (Z191100001119050),  and the Research Foundation of iFLYTEK, P.R. China.

{\small
	\bibliographystyle{ieee_fullname}
	\bibliography{reference}
}

\end{document}